\documentclass[journal]{IEEEtran}
\usepackage[left=0.625in,right=0.625in,top=0.75in,bottom=1in]{geometry}
\usepackage{ifpdf}
\usepackage{cite}
\usepackage{array}
\usepackage{dblfloatfix}
\usepackage{url}
\usepackage{amsmath}
\usepackage{ragged2e}
\usepackage{graphicx}
\usepackage{verbatim}
\usepackage{fancyhdr} % 添加页眉页脚
\usepackage[normalem]{ulem}
\usepackage{float}
\usepackage{lipsum}
\usepackage{multicol, blindtext}
\usepackage{caption}
\usepackage{subcaption}
\usepackage{xcolor}
\usepackage{setspace}
\definecolor{mypink2}{RGB}{0, 0, 255}
\definecolor{green}{RGB}{0, 128, 0}
\usepackage{multirow}
\usepackage{array}
\usepackage[utf8]{inputenc}   % not needed for LuaLaTeX/XeLaTeX
\usepackage[T1]{fontenc}
\usepackage{tabularx}
\usepackage[acronym]{glossaries}
\usepackage[linesnumbered]{algorithm2e}
\begin{document}

\title{Diffusion Models for Smarter UAVs: Decision-Making and Modeling}
\author{Yousef~Emami,~\IEEEmembership{Senior Member,~IEEE,}
        Hao~Zhou,~\IEEEmembership{Member,~IEEE,}\\
        Luis~Almeida,~\IEEEmembership{Senior Member,~IEEE,}
        and~Kai~Li,~\IEEEmembership{Senior Member,~IEEE}
\thanks{Yousef Emami is with Real-Time and Embedded Computing Systems Research Centre (CISTER), 4200-135 Porto, Portugal   (email:emami@isep.ipp.pt)}
\thanks{Hao Zhou is with the School of Computer Science, McGill University, Montreal, QC H3A 0E9, Canada. (email:hzhou098@uottawa.ca)}
\thanks{Luis Almeida is with CISTER and the Faculty of Engineering of the University of Porto (FEUP), 4200-465 Porto, Portugal (email: lda@fe.up.pt)}
\thanks{Kai Li is with the Interdisciplinary Centre for Security, Reliability and Trust (SnT), University of Luxembourg (email: kaili@ieee.org)}

\vspace{-20pt}}

\maketitle

\thispagestyle{fancy}            %更改plain状态，首页格式设为fancy
\chead{This paper has been accepted by IEEE Vehicular Technology Magazine. } 

\renewcommand{\headrulewidth}{1pt}      %把页眉线的宽度设为零，即去掉页眉线
\pagestyle{plain}

%TC:ignore
\begin{abstract}
Uncrewed Aerial Vehicles (UAVs) are increasingly used in modern communication networks. However, challenges in decision-making and digital modeling continue to hinder their rapid development. Reinforcement Learning (RL) algorithms face limitations such as low sample efficiency and limited data versatility, which are further amplified in UAV communications scenarios. Additionally, Digital Twin (DT) modeling presents significant challenges in decision-making and data management. RL models, often integrated into DT frameworks to address these issues, require large amounts of training data to make accurate predictions. Unlike traditional approaches that focus on class boundaries, Diffusion Models (DMs)—a new class of generative AI – learn the underlying probability distribution from training data and can generate reliable new patterns based on this learned distribution.
DT and RL have complementary roles in enabling intelligent, data-driven UAV operations. DMs further enhance this synergy by addressing data scarcity, improving modeling accuracy, and generating realistic scenarios, which benefit both DT simulations and RL training. In this paper, we explore the integration of DMs with RL and DT. Simulation results confirm the effectiveness and benefits of DMs in generating neighbor velocity estimates in a four-UAV swarm coordination task using Deep Reinforcement Learning (DRL).

\end{abstract}
%TC:endignore

\IEEEpeerreviewmaketitle

\section{Introduction}

Uncrewed Aerial Vehicles (UAVs) are increasingly used in sectors such as energy, public safety, agriculture, and smart cities, serving as data collectors, base stations, and relays. UAVs play a crucial role in developing 5th-generation (5G) networks, supporting the achievement of 5G goals, including Enhanced Mobile Broadband (eMBB), Ultra-Reliable and Low Latency Communications (URLLC), and Massive Machine-Type Communications (mMTC). UAVs are also expected to be pivotal in 6th-generation (6G) networks, enabling improved data collection and analysis. The main advantages of UAVs are rapid deployment, controlled mobility, and the ability to establish Line-of-Sight (LoS) communications, which support high-speed data transmission.
\par
Decision-making involves selecting the best course of action from multiple alternatives, such as determining optimal flight paths or scheduling data transmission. Along with digital modeling, it is an essential process in UAV communications. Reinforcement Learning (RL), a transformative approach within Artificial Intelligence (AI), offers promising solutions to complex decision-making challenges in UAV communications. RL can automate data collection tasks in UAV-assisted sensor networks; however, low sample efficiency often limits its effectiveness. Sample efficiency is the ability of an RL algorithm to learn effectively from a limited number of interactions with the environment. Moreover, its action-oriented nature requires prolonged and extensive interactions within dynamic environments, creating significant challenges for practical deployment. Immersive digital modeling systems, such as Digital Twins (DTs), replicate UAV elements, processes, dynamics, and firmware in a virtual counterpart. These physical and digital counterparts exchange inputs and operations seamlessly through real-time data communication. A UAV swarm with diverse aerial roles can be effectively modeled using DTs to enable collaboration, enhance safety, and address challenges such as the simulation-to-reality gap. However, DT modeling involves significant complexities in decision-making and data management. RL models, often used within DT frameworks, rely on extensive training data to develop accurate predictions. Acquiring sufficient and representative training data is a major challenge, particularly for systems with limited historical datasets or scenarios\cite{10246260}.

Diffusion Models (DMs), a class of generative AI, show significant potential in addressing the challenges mentioned above. Unlike traditional approaches that focus on class boundaries, DMs learn the underlying probability distribution from training data, allowing them to generate reliable new samples based on this learned distribution. DMs can synthesize data to help RL models overcome low sample efficiency, improve their policies for better handling of UAV dynamic environments, and create realistic simulation environments for UAV training tasks. Additionally, DMs can help DT-assisted UAVs overcome data scarcity, improve decision-making when integrated with RL, and enhance UAV modeling accuracy. Sun et al. \cite{sun2024generative} demonstrate the use of generative AI to enhance RL. Moreover, Zhu et al. \cite{zhu2023diffusion} survey the applications of DMs in RL, and Sun et al. \cite{sun2024generative2} systematically showcase the adoption of generative AI in optimizing UAV communication and networking problems.
\par
\begin{figure} [h] 
        \centering
%        \captionsetup{justification=raggedright}
%        \includegraphics[ width=3.50in, height =3in]{Drawing1.png}
        \includegraphics[ width=0.45\textwidth]{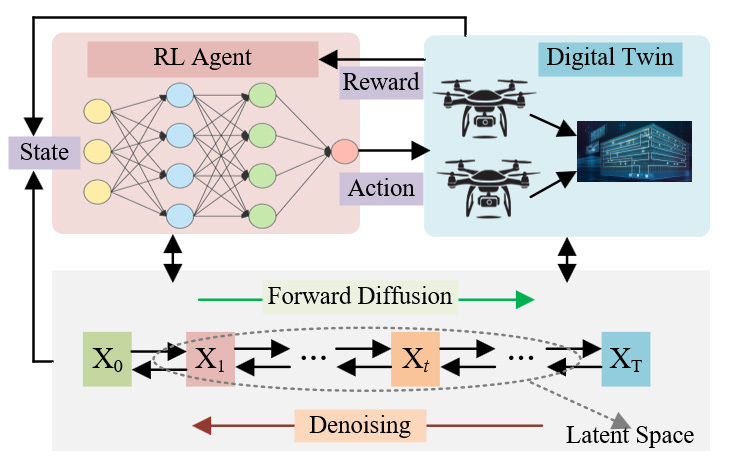}
    	\caption{An overview of DMs integrating with DT and RL technologies for UAV communications where DMs benefit RL with synthetic data and improved policy network, and benefit DT with synthetic data and dynamic modeling.}
    	\label{fig:digital}
        \vspace{-10pt}
\end{figure}

DT and RL have complementary roles in enabling intelligent, data-driven UAV operations. 
DMs further enhance this synergy by addressing data scarcity, improving modeling accuracy, 
and generating realistic scenarios, benefiting both DT simulations and RL training. 
In particular, this work provides an integrative perspective on how DMs can support RL-assisted UAV communications by improving sample efficiency, enriching training data, and enabling realistic training environments. 
Rather than providing an exhaustive survey of the rapidly growing literature, our focus is on organizing representative DM--RL--DT integration directions and illustrating their relevance to UAV systems.
Fig.~\ref{fig:digital} shows the integration of DMs with RL and DT for UAV systems.

The core contributions of this work are twofold. 
First, we provide a structured and integrative perspective on the role of DMs in UAV communications by connecting representative diffusion-augmented RL and diffusion-assisted DT applications within a common framework. 
Rather than considering these components separately, we illustrate how different DM variants can support representative UAV communication and networking tasks, and how these integration directions can jointly address recurring challenges such as data scarcity, low sample efficiency, and dynamic modeling complexity.
Second, we complement this integrative analysis with an illustrative and representative simulation-based case study on a 4-UAV swarm coordination using Deep Reinforcement Learning (DRL). We quantitatively compare the performance of the system when using DMs, Generative Adversarial Networks (GANs), and Variational Autoencoders (VAEs) under identical training settings. 
The results show that DMs exhibit more stable learning dynamics with lower performance variance, providing empirical support for their suitability as generative components in adaptive and safety-sensitive UAV systems. 
Instead of serving as an exhaustive benchmark of diffusion-based UAV systems, the case study is intended to provide a concrete illustration of one DM-assisted RL integration scenario and to complement the broader conceptual discussion. 
Together, the integrative discussion and the representative case study provide conceptual guidance and practical insight into the potential use of DMs in RL- and DT-assisted UAV systems.

The rest of this paper is organized as follows: Section \ref{sec2} provides an overview of DMs, UAV communications, and their implementation challenges. Section \ref{sec3} presents DM integration with RL for UAV communications and provides a case study on synthetic velocities for UAV swarm. Section \ref{sec4} presents DM integration with DT for UAV communications. Section \ref{sec5} outlines future work and open issues. Finally, Section \ref{sec6} concludes the paper.

\section{Overview of Diffusion Model and UAV Communications} \label{sec2}

This section provides an overview of DMs, UAV communications, and implementing DMs for UAV applications.

\subsection{Diffusion Model}

DMs are a class of probabilistic generative models that incrementally corrupt data by adding noise and then learn to reverse this process to generate samples. DMs are primarily used in two application paradigms: unconditional and conditional generation, similar to other generative models such as VAEs, GANs, and flow-based models. Unconditional generation evaluates the fundamental capabilities of the model, while conditional generation targets application-specific objectives by enabling control over generated outputs.
\par
A conditional DM is a class of generative models that extends the standard (unconditional) diffusion framework by incorporating external information (conditions) to guide the generative process. Formally, while an unconditional DM learns the data distribution $p(x)$ via a denoising diffusion process, a conditional DM learns the conditional distribution $p(x|c)$  where c is a condition such as a class label, text prompt, image, semantic map, or other structured data. A conditional DM has three main components: the forward process, the reverse process, and the training objective. The forward process adds Gaussian noise to clean data over time, independent of any condition. The reverse process models $p_\theta(x_{t-1} | x_t, c)$, using a neural network that predicts the noise or denoised sample based on the noisy input $x_t$ and condition $c$, integrated via mechanisms like concatenation, cross-attention, gradient guidance, or adaptive normalization. The model is trained by minimizing the error between the true and predicted noise, conditioned on $c$.
\par
Conditional DMs are highly valuable in UAV communications because they enable precise control over outputs based on user inputs. By conditioning variables such as labels, text, or images, DMs allow users to guide the generation process to achieve desired outcomes. This capability is especially important for creative and operational tasks, where managing the style, class, or attributes of generated content, such as specific UAV states, is essential. In UAV communications, conditional generation enables fine-tuned control over state outputs, ensuring the system meets precise operational demands. 
\par
Denoising Diffusion Probabilistic Models (DDPMs) use two Markov chains: a forward chain that gradually corrupts the data by adding noise, and a reverse chain that reconstructs the original data. The forward chain maps the data distribution to a simple prior, typically a standard Gaussian distribution. The reverse chain is trained with Deep Neural Networks (DNNs) to approximate the transition kernels and reverse the noise-adding process. New data points are generated by sampling from the prior distribution and then sequentially passing through the reverse Markov chain to produce realistic outputs.
\par
Score-based Generative Models (SGMs) work by perturbing data with progressively increasing levels of Gaussian noise and using a DNN to estimate the score functions of the resulting noisy data distributions, conditioned on the noise levels. New samples are generated by integrating these score functions at decreasing noise levels using various sampling strategies, such as Langevin Monte Carlo, stochastic differential equations, ordinary differential equations, or hybrid approaches. A key advantage of SGMs is the decoupling of the training and sampling processes, which allows different sampling techniques to be applied after the score functions have been estimated.
\par
DDPMs, SGMs, and conditional DMs are closely related approaches within diffusion-based generative modeling. Both DDPMs and SGMs add noise in a forward process and learn to reverse it, but differ in training: DDPMs optimize a variational bound, while SGMs estimate score functions. However, their objectives can be equivalent under certain conditions. Conditional DMs extend these models by integrating external inputs (such as labels or prompts) through mechanisms like cross-attention or classifier guidance. The choice between DDPMs and SGMs depends on application needs: DDPMs are simple and stable, making them popular for tasks like synthetic data generation, while SGMs offer flexible sampling for complex tasks. In summary, model selection should balance trade-offs in controllability, sampling efficiency, and generation quality, depending on the use case\cite{10.1145/3626235}.

%TC:ignore
\begin{table*}[ht]
\centering
\caption{Categorization of Diffusion Model applications for enhancing Reinforcement Learning in UAV communications across three key aspects: synthetic data generation, policy network improvement, and realistic training environment simulation.}
\label{dm_uav}
%\setstretch{1.05}
\resizebox{1\textwidth}{!}{%
\begin{tabular}{|m{2cm}|m{7cm}|m{7cm}|m{3cm}|m{1.5cm}|}
\hline
\textbf{Aspect} & \textbf{Description} & \textbf{Key Benefits} & \textbf{Application Context} & \textbf{References} \\ \hline
\textbf{Generating Synthetic Data} 
& DMs generate diverse synthetic data for RL training, addressing data scarcity in UAV communications. 
& - Enhanced training diversity \newline - Alignment with environmental dynamics \newline - Novel velocity samples for UAV swarms 
& Trajectory planning and collision avoidance in UAV swarms 
& \cite{zhu2023diffusion} \\ \hline

\textbf{Improved Policy Network} 
& DMs improve RL policy networks (e.g., SAC, TD3) by generating action distributions from observed states. 
& - Better decision-making \newline - Exploration of diverse actions \newline - Adaptability in dynamic environments 
& Dynamic trajectory optimization and adaptive action selection 
& \cite{du2024diffusion}, \cite{tong2024diffusion} \\ \hline

\textbf{Training Environment} 
& DMs simulate realistic environments for UAV training in high-risk scenarios such as Search-and-Rescue (SAR), where real-world trial-and-error training is impractical. 
& - Risk-free training \newline - Optimization for real-world missions \newline - Modeled complex environments 
& SAR missions, hazardous environments 
& \cite{sun2024generative} \\ \hline
\end{tabular}}
\vspace{-10pt}
\label{tab:dm_uav}
\end{table*}
%TC:endignore

\subsection{UAV Communications}

UAVs have become integral to modern technological advancements, driving transformative changes across multiple sectors. They have revolutionized applications in agriculture, public safety, environmental monitoring, and security. In agriculture, UAVs are especially promising for enabling precision farming, aligning with the European Union's focus on sustainable and eco-friendly agricultural practices, where reliable wireless communication supports large-scale sensing and timely data aggregation. Moreover, UAVs have shown exceptional utility in evaluating hazardous environments, conducting SAR operations, collecting evidence for investigations, and identifying potential threats, all of which critically depend on robust air-to-ground and air-to-air communication links for coordination and situational awareness.
\par
UAV communications differ fundamentally from terrestrial networks due to high altitude and mobility, frequent LoS links with ground sensors, heterogeneous Quality of Service (QoS) requirements, strict Size, Weight, and Power (SWAP) constraints, and the need to jointly optimize mobility control with communication scheduling and resource allocation. Unlike terrestrial systems, UAV networks tightly couple mobility decisions with communication performance. This coupling is further amplified by 3D mobility, LoS maintenance, energy constraints, and highly dynamic environments, where mobility continuously reshapes network topology, link availability, and end-to-end reliability.
\par
UAVs often serve as aerial data collectors and communication relays at much higher altitudes than terrestrial gateways. While terrestrial gateways are typically deployed at 10--25 meters, UAVs can operate at altitudes up to 122 meters under regulatory limits. Such altitude flexibility directly affects air-to-ground channel characteristics and communication quality, enabling wider coverage but also altering link budgets and coverage--capacity trade-offs. Unlike terrestrial channels dominated by shadowing and multipath, UAV-to-ground links are largely LoS, resulting in higher channel gains and more reliable connectivity that is sensitive to altitude, positioning, and trajectory design.
UAV mobility further introduces time-varying air-to-ground channels, leading to rapid fluctuations in link quality if not properly managed. At the same time, controlled mobility enables communication-aware trajectory adaptation to improve network performance, explicitly aligning motion planning with link stability, connectivity duration, and QoS objectives. However, strict SWAP constraints limit endurance and onboard processing, requiring tight coordination among communication reliability, energy efficiency, and mobility control.

\subsection{Implementing DMs for UAV applications}
Two significant technical hurdles arise when implementing DMs for UAV applications. First, training instability often results from suboptimal hyperparameter selection or poorly configured noise schedules. This can manifest in several ways: The U-net architecture commonly used in DMs requires precise calibration of layer depths, attention mechanisms, and normalization parameters to avoid gradient-related issues. Additionally, an incorrectly set learning rate can cause erratic loss fluctuations that prevent the model from converging and reduce training effectiveness.
\par
Second, while DMs show greater robustness to mode collapse compared to GANs, they are still susceptible to producing repetitive or minimally varied results when trained on insufficiently diverse datasets. This limitation poses a particular challenge for UAV applications, where the system must handle a wide range of possible scenarios and environmental conditions.
The output quality of the model directly correlates with the extent and representativeness of its training data, which is particularly important for UAV operations that must remain reliable across diverse real-world conditions. 
In this work, DMs are trained in a centralized, pre-deployment setting and serve as generative support components for data synthesis, environment modeling, and RL policy learning rather than as real-time UAV control modules. Thus, DM-assisted information generation is part of the model-development and policy-learning process and is not performed within the real-time physical UAV control loop. This separation allows computationally intensive generative modeling to be carried out with sufficient resources, while the resulting trained control policy is subsequently used for real-time UAV deployment.

\section{Diffusion Model Integration with RL} \label{sec3}

\begin{figure} [h] 
        \centering
        \includegraphics[width=0.45\textwidth]{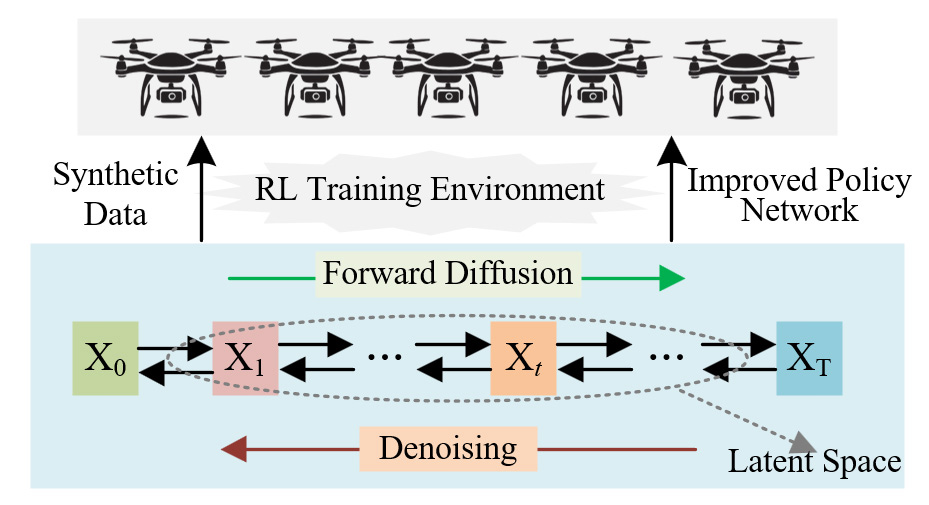}
    	\caption{An overview of DM integrating with RL technologies for UAV communications where DM benefits RL with synthetic data training environment.}
    	\label{fig:digital3}
\end{figure}

This section investigates how DM integration with RL contributes to generating synthetic data, creating simulation environments, and improving the policy network for better decision-making.
A well-known limitation of RL algorithms is low sample efficiency, and data can be scarce and limited in UAV communications. Data collection with RL can be costly or impractical in real-world UAV scenarios, as learning optimal policies often requires prolonged and extensive environmental interactions. Moreover, generalization capabilities are limited. Fig.~\ref{fig:digital3} shows how DMs can enhance decision-making capabilities with RL for UAV communications,  providing UAVs with synthetic data and improved simulated environments.

\subsection{Generating Synthetic Data}

A common use of DMs in UAVs is to generate synthetic data, addressing the data scarcity often encountered in UAV communications. DMs act as natural data synthesizers for RL training, producing diverse and realistic data. Traditional data augmentation in RL typically involves adding minor perturbations to states and actions to maintain consistency with environmental dynamics. However, DMs can generate more varied synthetic samples, enhancing training diversity while closely aligning with the environment's underlying dynamics.
\par
In a UAV swarm, inter-UAV state information can become incomplete or degraded as mobility changes the communication conditions among neighboring UAVs. In particular, increasing inter-UAV separation can reduce link quality and consequently affect the reliability of exchanged or perceived neighbor-state information. This makes coordinated decision-making more challenging when each UAV relies primarily on local observations. DMs can help mitigate this limitation by generating plausible synthetic neighbor-velocity information from learned state distributions. The main idea is to generate velocity states that resemble those expected from neighboring UAVs and use them to augment the information available to the RL agent during policy learning \cite{10610665}.

DMs use advanced sampling strategies to optimize output diversity and fidelity, which are crucial for UAV decision-making and modeling. For example, Ancestral Sampling incorporates prior domain knowledge to guide the reverse diffusion process, significantly improving sample realism. Additionally, sample quality can be enhanced by iteratively refining outputs with domain-specific constraints, ensuring the generated data adheres to the physical and operational boundaries relevant to UAV systems. These techniques collectively contribute to more reliable and applicable synthetic data generation for UAV applications.

\begin{figure} 
    \centering
    \includegraphics[width=0.48\textwidth]{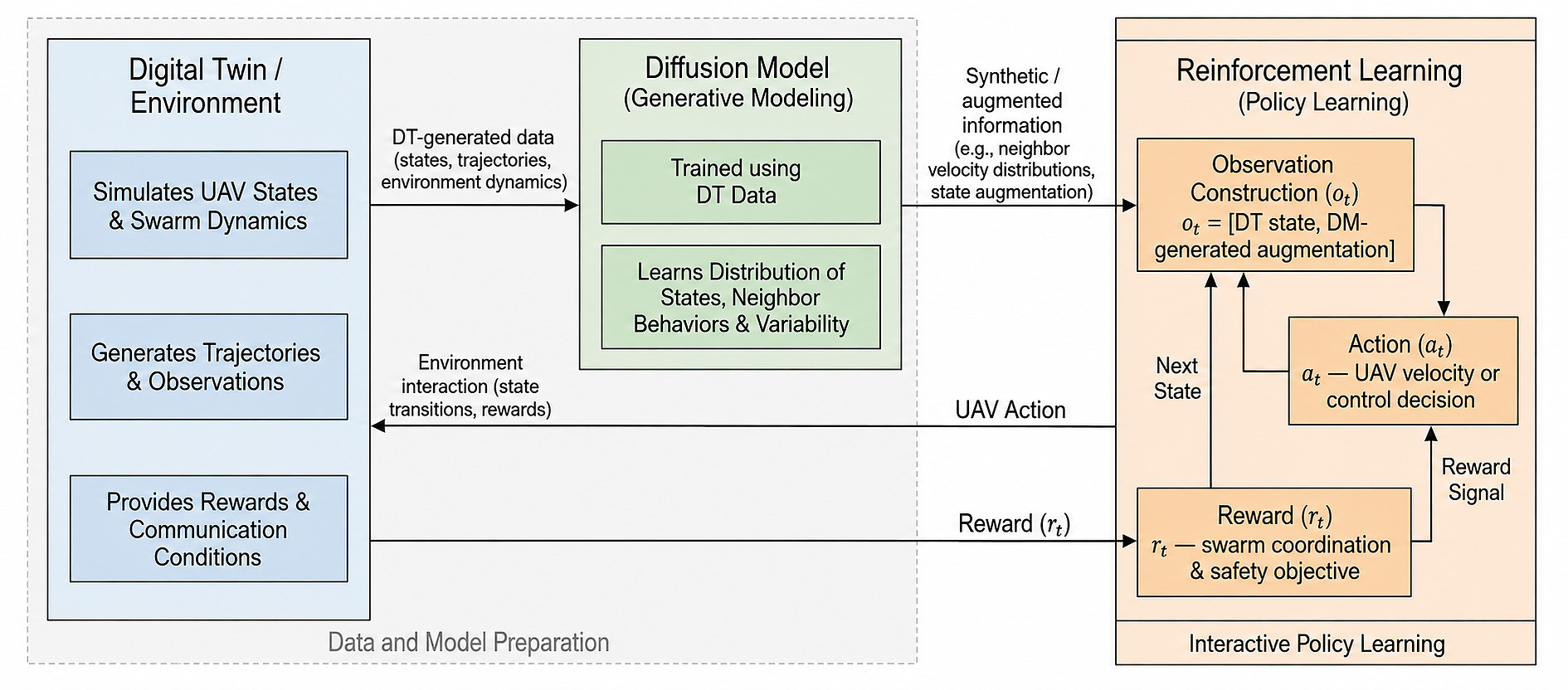}
    \caption{End-to-end pipeline illustrating the synergy between DT, DM, and RL for UAV swarm coordination.  The DT provides the simulated environment, system observations, and data used for model preparation.  The DM is trained using DT-generated data to learn the distribution of UAV states and neighboring behaviors, and provides synthetic or augmented information (e.g., neighbor velocity estimates) to enrich the RL agent's observations. 
During interactive policy learning in the DT environment, the RL agent selects control actions based on the augmented observations, receives rewards and subsequent states from the DT, and iteratively improves its policy for coordinated swarm behavior.}
    \label{drl1011}
\end{figure}

\begin{figure} 
    \centering
    \includegraphics[width=0.45\textwidth]{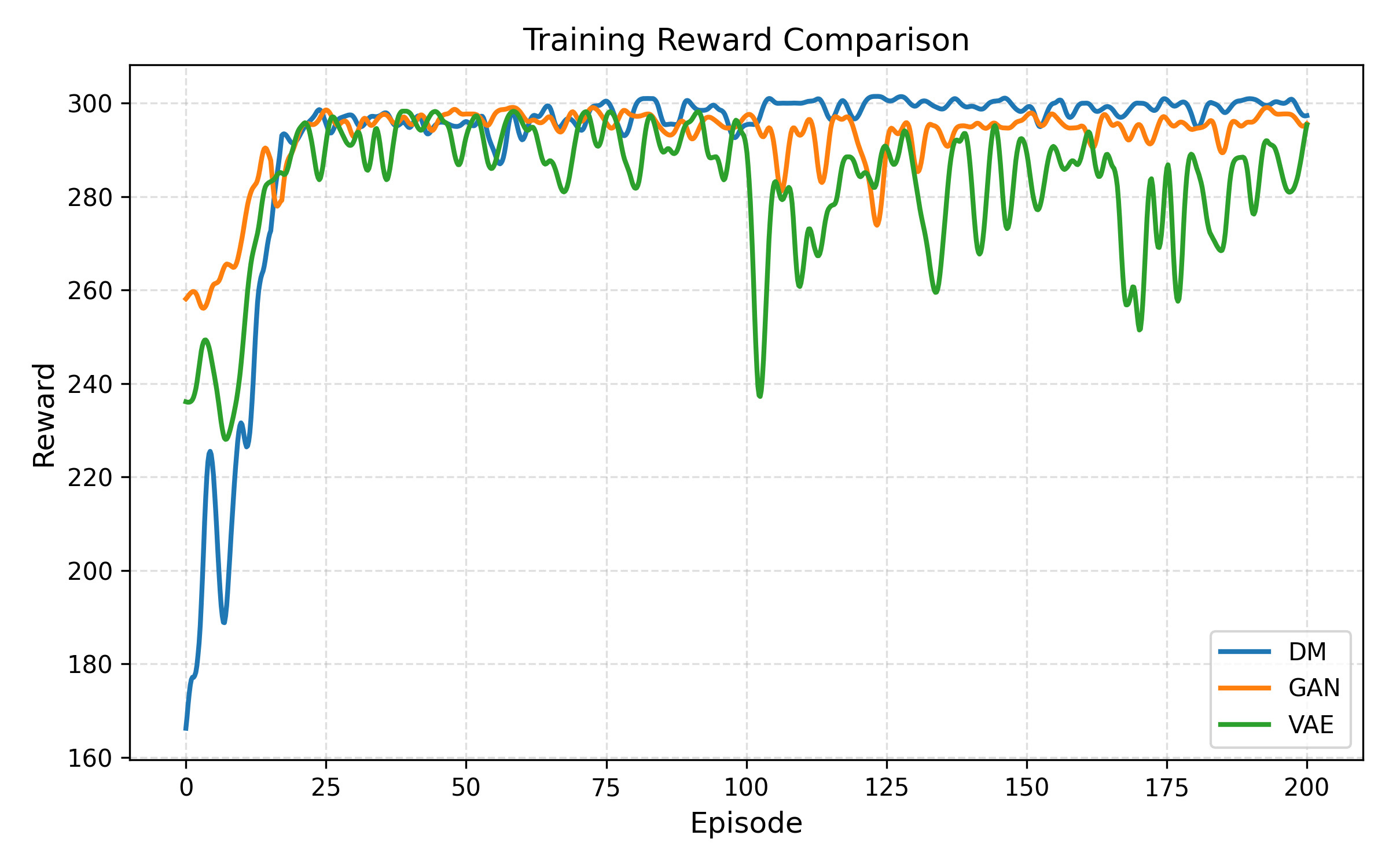}
    \caption{Average reward per episode for a 4-UAV swarm coordination task across 200 training episodes, comparing DMs, GANs, and VAEs. DMs achieve better training stability.}
    \label{drl1010}
\end{figure}

Fig.~\ref{drl1011} illustrates the overall system architecture and data flow of the proposed framework, highlighting the interactions among the DT, DM, and RL components. The DT acts as a shared environment and data source, simulating UAV states, swarm dynamics, and communication conditions, and generating trajectories, observations, and rewards. These DT-generated data are used for model preparation, through which the DM learns the distributions of system states and neighboring UAV behaviors. During interactive policy learning in the DT environment, the RL agent operates in a standard sequential decision-making loop. The DM provides synthetic or augmented information to enrich the agent's observations, while the DT executes actions, returns rewards, and provides subsequent states for policy improvement. This design distinguishes data and model preparation from interactive policy learning, while both stages remain part of the pre-deployment training process prior to real-time UAV control.

To assess the benefits of DM when compared to alternatives, namely GANs and VAEs, in supporting RL-based systems, we resort to the referred 4-UAV coordination case study in which each agent adjusts its velocity based on both its own observations and synthetically generated neighbor velocities.

The four UAVs operate within a $200 \times 200$~m area. At each 1-s step, each UAV updates its position according to its selected speed and a stochastic heading variation uniformly sampled between $-0.2$ and $0.2$~rad. The resulting inter-UAV distances are used to determine pairwise UAV-to-UAV communication conditions through a free-space path-loss (FSPL) model at 2.4~GHz. As the UAVs move, their relative positions therefore affect the quality of the information available from neighboring UAVs. In particular, weaker links result in more strongly attenuated and noisier generative-model-synthesized neighbor-state estimates, which are subsequently incorporated into the RL observations.

Specifically, the DM is implemented as a denoising diffusion probabilistic model (DDPM) with 50 diffusion steps and a linear noise schedule ranging from $10^{-4}$ to $0.02$. Its MLP-based noise-prediction network has a hidden width of 256 and uses the conditioning context, noisy target, and timestep embedding as inputs. The DDPM is pretrained for 10 epochs on 8,000 synthetic velocity samples using Adam with a learning rate of $10^{-3}$ and a batch size of 64, after which its parameters are fixed during policy learning.
The GAN employed a generator with layers [64, 128] and a discriminator with layers [128, 64], and the VAE used a symmetric encoder--decoder architecture with layers [128, 64] and [64, 128], ensuring comparable model capacity across all methods.

Each UAV is controlled by an independent DQN agent without policy or parameter sharing. The action space consists of 15 discrete speed levels ranging from 1 to 15~m/s. The observation $o_t$ is a 16-dimensional vector consisting of the UAV's own 4-dimensional velocity descriptor and generative-model-synthesized estimates of the three neighboring UAVs' 4-dimensional velocity descriptors. These neighbor estimates incorporate the corresponding communication impairments rather than directly using the ground-truth neighbor velocities.

The reward is defined according to the physical separation among UAVs: for each neighboring UAV located more than the predefined safety distance of 15~m away, the agent receives one reward unit. The resulting reward therefore ranges from 0 to 3 at each step and encourages the UAVs to maintain sufficient inter-agent separation

The DQN uses a two-hidden-layer MLP with 128 units per layer, a learning rate of $4\times10^{-5}$, a discount factor of 0.99, a replay-buffer capacity of 20,000 transitions, and a batch size of 128. Learning starts after 1,000 transitions, and the target network is updated every 500 gradient steps. Training consists of 200 episodes with 100 steps per episode. UAV positions are randomly re-initialized and velocities are reset to 1~m/s at the beginning of each episode. The reported results correspond to 5 independent training runs. 
Accordingly, the case study evaluates UAV coordination under distance-dependent, communication-impaired neighbor information rather than under perfect inter-UAV state availability. Its objective is to assess the effect of generative observation augmentation on communication-aware swarm coordination, rather than to provide a complete packet- or QoS-level evaluation of the wireless network.

The results in Fig.~\ref{drl1010} show that DMs and GANs achieve comparable asymptotic performance, both reaching high reward levels close to 300 with similar convergence behavior.
When compared to VAEs, DMs maintain consistently high rewards after convergence, while VAEs exhibit noticeably larger performance degradations and fluctuations during mid- and late-stage training.

The DM's iterative denoising process provides stable neighbor-velocity estimates throughout the reported training trajectory. While the GAN reaches a broadly comparable final reward level, the DM exhibits more consistent episode-to-episode learning behavior, whereas the VAE shows larger reward fluctuations during the mid- and late-training stages.

\subsection{Improved Policy Network}

In this work, the policy network refers to the component of an RL agent that maps observed states to an action distribution, from which control actions are sampled. 
The reverse process of DMs is well-suited to this role, as its multi-round denoising iterations enable the modeling of complex and potentially multi-modal action distributions beyond standard parametric policies. DM-based Soft Actor-Critic (SAC) \cite{du2024diffusion} schemes can enhance UAV adaptability to dynamic environments by replacing conventional unimodal action parameterizations with diffusion-generated distributions conditioned on observed states. 
The DMs can be used directly as the policy network in RL algorithms such as SAC, enabling the generation of action distributions based on observed states rather than fixed-form assumptions. 
In this context, policy network improvement refers to increased expressiveness and diversity of the action distribution, rather than architectural complexity.
This capability enhances decision-making by allowing UAV agents to sample actions that better reflect environmental uncertainty and dynamic interactions in complex scenarios.
Similarly, DMs can be integrated into the actor network of the conventional TD3 algorithm~\cite{tong2024diffusion}, where they provide a more diverse set of candidate actions through the learned reverse diffusion process. 
Here, enhanced decision-making denotes improved action selection quality under partial or noisy observations, achieved by sampling from a richer generative action space.

\subsection{Training Environment}

Because DMs excel at learning and modeling complex data distributions, they are particularly well-suited for simulating realistic environments for UAV training tasks. In scenarios such as SAR operations, a UAV cannot train through trial-and-error. In these situations, the risk of damaging equipment or injuring people makes real-time trial-and-error training impractical. Instead, UAVs can be trained in simulated environments to ensure safety while optimizing performance for real-world missions \cite{sun2024generative}. Table \ref{dm_uav} provides a summary of the discussed content.

\subsection{Lesson Learned}

To effectively facilitate the operation of UAV swarms, there is an urgent need for precise modeling and control. However, existing technologies fail to fulfill these objectives, particularly when communication dynamics and channel uncertainty are tightly coupled with swarm behavior. 
Generative AI provides a powerful approach for modeling the complex and dynamic Air-to-Ground (A2G) communication environment, which is influenced by UAV altitude, orientation, and urban obstacles. Unlike conventional methods, generative AI captures the joint distributions of physical-layer parameters, enabling accurate estimation of channel conditions and communication performance. Implementations include GAN-based models for high-speed UAV channel estimation, two-stage VAE frameworks for adaptive path parameter generation, and DMs integrated with UAVs for enhanced semantic communication. These approaches improve key QoS metrics such as latency, throughput, and reliability in challenging A2G scenarios \cite{zhao2025generative}.

\section{Diffusion Model Integration with Digital Twin} \label{sec4}

This section investigates how DM integration with DT contributes to data transmission and synchronization, data scarcity, conditional generation, task coordination, and dynamic modeling as depicted in Fig. \ref{fig:digital3}.

%TC:ignore
\begin{table*}[htb]
\centering
\caption{Diffusion Model Applications in Digital Twin-Assisted UAV Systems for Communication Enhancement, Task Coordination, and Dynamic Control}
\label{dm_dt_uav}
\resizebox{1\textwidth}{!}{%
\begin{tabular}{|m{2cm}|m{7.5cm}|m{6cm}|m{3.5cm}|m{1.5cm}|}
\hline
\textbf{Aspect} & \textbf{Description} & \textbf{Key Benefits} & \textbf{Application Context} & \textbf{References} \\ \hline
\textbf{Data Transmission and Synchronization} 
& Integration of DMs into DT frameworks to support data consistency and synchronization, enabling generative modeling that improves robustness and alignment of DT updates under noisy, incomplete, or dynamic conditions. 
& - Real-time synchronization \newline - Enhanced communication reliability \newline - Seamless coordination in dynamic environments 
& UAV-DT communication in dynamic and noisy environments 
& \cite{tao2024wireless} \\ \hline
\textbf{Data Scarcity} 
& Combination of DM and DT to generate diverse synthetic datasets reflecting real-world dynamics, addressing data scarcity for RL training. 
& - Enhanced training data availability \newline - Improved representation of real-world scenarios 
& RL model training for UAVs in data-limited scenarios 
& \cite{10246260} \\ \hline
\textbf{Conditional Generation} 
& DMs enable conditional generation and decision-making in DT environments. Return-conditioned DMs generate solutions optimized for specific objectives. 
& - Efficient training \newline - Enhanced inference performance \newline - Optimization of decision-making 
& Network decision-making and optimization in DT-assisted UAVs 
& \cite{yu2024regularized} \\ \hline
\textbf{Task Coordination} 
& Return-conditioned DMs facilitate planning and trajectory generation for multi-task operations, leveraging advanced DT simulation and modeling capabilities. 
& - Favorable multi-task performance \newline - Optimized task allocation 
& Multi-task UAV operations with complex interaction requirements 
& \cite{10246260} \\ \hline
\textbf{Dynamic Modeling} 
& DroneDiffusion framework leverages DDPM to model and control UAV dynamics under multimodal and stochastic conditions. Noise and denoising steps capture and predict state uncertainties. 
& - Improved control adaptability \newline - Robust trajectory tracking \newline - Generalization across dynamic conditions 
& Precise trajectory control in unpredictable environments 
& \cite{das2024dronediffusion} \\ \hline
\end{tabular}}
\label{tab:dm_dt_uav}
\end{table*}
%TC:endignore

\subsection{Data Transmission and Synchronization}

DMs can be integrated into a DT framework to support data consistency and synchronization between physical UAV systems and their virtual counterparts.
Specifically, DMs are trained offline on DT-generated or DT-collected data to learn the underlying distribution of system states, environmental dynamics, and communication-relevant conditions.
By capturing these distributions, DMs can generate realistic and coherent state realizations that help maintain alignment between the DT and the physical UAV under dynamic and uncertain operating conditions.
In this context, the role of DMs is to enhance robustness and consistency of DT-assisted communication and synchronization by providing distribution-aware generative modeling~\cite{tao2024wireless}.

\subsection{Data Scarcity}

RL models rely on training data to learn and generate accurate predictions. However, obtaining adequate and representative training datasets can be challenging, especially in scenarios with limited historical data. To address data scarcity, operators may use strategies such as data augmentation or generating synthetic datasets to enhance the availability and diversity of training data \cite{10246260}. Combining DM and DT enables the creation of diverse synthetic data that accurately reflects real-world dynamics, which can be used to train RL models more effectively.

\subsection{Conditional Generation}

ML models support decision-making, maintenance strategies, process optimization, and resource allocation in DT environments by analyzing data and extracting insights from observed trends. DTs collect network data, recreate the network environment, and provide a foundation for training ML models. RL models enhance this process by offering data-driven solutions for network decision-making, using real-time network conditions and historical data to optimize performance. A DM enables conditional generation and network decision-making through reverse diffusion and Markovian transition processes. A return-conditional DM guides solution generation based on specific returns. The model learns to make decisions that optimize particular objectives by conditioning on target returns. A return-conditional DM can significantly improve the effectiveness of a DT-assisted UAV by increasing training efficiency and inference performance\cite{yu2024regularized}.

\subsection{Task Coordination}

Task coordination for UAVs often involves executing multiple tasks concurrently, collaborating with other UAVs, and interacting with environmental objects. DTs designed to address these complexities must incorporate advanced modeling and simulation capabilities, as well as seamless integration of ML techniques, to effectively manage interactions and optimize performance. Utilizing return-conditioned DM for planning or trajectory generation achieves favorable performance in multi-task operations but depends on well-defined reward functions, which require significant human effort.

\subsection{Dynamic Modeling}

The DroneDiffusion framework \cite{das2024dronediffusion} uses DDPM to learn complex quadrotor dynamics. It aims to improve quadrotor control by accurately modeling the multimodal and stochastic nature of flight dynamics, including unpredictable environmental factors such as wind or payload changes. The DDPM operates through a sequence of noise addition and denoising steps. In the forward process, noise is added to data representing quadrotor dynamics, resulting in multiple variations that capture possible state uncertainties. The model then gradually denoises these states in the reverse process and learns a predictive distribution of the quadrotor dynamics. This denoised prediction of the residual dynamics is integrated into a controller, which uses it to make robust adjustments for stable and precise trajectory tracking in real time. The framework has shown enhanced stability and adaptability in both simulated and real-world tests across different trajectories and conditions. Table \ref{dm_dt_uav} provides a summary of the discussed content.

\begin{comment}
\begin{table}[ht]
\caption{Applications of DMs for Digital Modeling in UAV communications}
\label{dm_dt_uav}
\centering
\begin{tabular}{|m{0.5cm}|m{2.5cm}|m{4cm}|}
\hline
\textbf{ID} & \textbf{Application} & \textbf{Description} \\ \hline
\cite{tao2024wireless} & Data Transmission and Synchronization & DMs integrated into DTs enhance UAV data transmission by performing tasks such as compression, noise reduction, and signal reconstruction, ensuring efficient synchronization. \\ \hline
\cite{emami2024use} & Data Scarcity & Combining DM and DT enables the creation of synthetic data reflecting real-world dynamics, effectively addressing data scarcity for RL training. \\ \hline
\cite{huang2024digital}, \cite{yu2024regularized} & Conditional Generation & Return-conditional DMs in DTs guide solution generation for optimizing specific objectives, enhancing DT-assisted UAV training and decision-making efficiency. \\ \hline
\cite{emami2024use} & Task Coordination & DTs, integrated with return-conditioned DMs, optimize multi-task operations for UAVs by enabling advanced planning and trajectory generation based on well-defined rewards. \\ \hline
\cite{das2024dronediffusion} & Dynamic Modeling & The DroneDiffusion framework uses DDPM to model complex quadrotor dynamics, improving control and stability by simulating multimodal, stochastic flight scenarios. \\ \hline
\end{tabular}
\end{table}
\end{comment}

\subsection{Practical Validation of DT Frameworks}
Gurses et al. \cite{gurses2404digital} conducted a detailed case study on AI-assisted localization of signal sources within the DT environment of the AERPAW platform.
The experimental results compare three scenarios: development solely in the DT environment; development in DT and testing in the real world without calibration; and development in DT with calibration using real-world data, followed by testing in the real world. The results showed significant performance improvements from real-world calibration, demonstrating the effectiveness of the DT framework in bridging the gap between simulation and reality. This aligns with our exploration of DMs as a means to improve DT-based UAV communications. DMs help generate realistic synthetic data, improve dynamic modeling, and enable robust task coordination, thereby strengthening the calibration process and enhancing overall DT performance in real-world UAV applications.

%\section{Future Works and Open Issues} \label{sec5}
\section{Most Recent Trends} \label{sec5}

Recent research has explored several promising directions for applying DMs to UAV communications. One direction is to use DMs as network optimizers. By learning the distribution of high-quality solutions and repeatedly sampling from feasible regions, DMs can better capture global solution structures and support optimization in high-dimensional UAV scenarios \cite{liang2024diffusion}.

DMs also show potential for improving UAV security through anomaly and intrusion detection. Their ability to model complex data distributions makes them suitable for multimodal and varying operating conditions. In particular, DDPMs can support unsupervised intruder detection through reconstruction-error-based methods, reducing reliance on labeled anomaly data \cite{CAI2025107064}.

Another direction concerns improving decision-making beyond conventional MLP-based DRL architectures. MLPs can struggle with dynamic state spaces and conflicting objectives, whereas DMs can capture richer dependencies among environmental states and potentially support more balanced decisions in uncertain UAV environments.

Finally, improving the efficiency and scalability of DMs remains important for practical deployment. Advances in accelerated sampling, latent-space modeling, and hardware-aware execution are expanding their applicability. Future UAV systems may therefore adopt hybrid architectures in which DMs support training, modeling, and decision support, while lightweight controllers handle time-critical onboard execution.

\section{Conclusion} \label{sec6}

This paper explores the integration of DMs with RL and DT for UAV communications. DMs can support RL by alleviating data scarcity, improving sample efficiency, and enhancing policy learning, while also benefiting DT through synthetic data generation, synchronization, task coordination, and dynamic modeling. Simulation results further demonstrate the effectiveness of DMs for neighbor-velocity generation in a 4-UAV DRL coordination task. Future work will investigate DMs for UAV network optimization and security.

%\section*{Acknowledgment}

\bibliographystyle{IEEEtran}
\bibliography{references}

@article{zhu2023diffusion,
  title={Diffusion models for reinforcement learning: A survey},
  author={Zhu, Zhengbang and Zhao, Hanye and He, Haoran and Zhong, Yichao and Zhang, Shenyu and Guo, Haoquan and Chen, Tingting and Zhang, Weinan},
  journal={arXiv preprint arXiv:2311.01223},
  year={2023}
}

@article{tao2024wireless,
  title={Wireless network digital twin for 6g: Generative ai as a key enabler},
  author={Tao, Zhenyu and Xu, Wei and Huang, Yongming and Wang, Xiaoyun and You, Xiaohu},
  journal={IEEE Wireless Communications},
  volume={31},
  number={4},
  pages={24--31},
  year={2024},
  publisher={IEEE},
  month={Aug.}
}

@ARTICLE{sun2024generative,
  author={Sun, Geng and Xie, Wenwen and Niyato, Dusit and Mei, Fang and Kang, Jiawen and Du, Hongyang and Mao, Shiwen},
  journal={IEEE Wireless Communications}, 
  title={Generative AI for Deep Reinforcement Learning: Framework, Analysis, and Use Cases}, 
  year={2025},
  volume={32},
  number={3},
  pages={186-195},
  month={Jan.}
  }

@ARTICLE{du2024diffusion,
  author={Du, Hongyang and Li, Zonghang and Niyato, Dusit and Kang, Jiawen and Xiong, Zehui and Huang, Huawei and Mao, Shiwen},
  journal={IEEE Transactions on Mobile Computing}, 
  title={Diffusion-Based Reinforcement Learning for Edge-Enabled AI-Generated Content Services}, 
  year={2024},
  volume={23},
  number={9},
  pages={8902-8918},
  month={Jan.}}

@ARTICLE{sun2024generative2,
  author={Sun, Geng and Xie, Wenwen and Niyato, Dusit and Du, Hongyang and Kang, Jiawen and Wu, Jing and Sun, Sumei and Zhang, Ping},
  journal={IEEE Network}, 
  title={Generative AI for Advanced UAV Networking}, 
  year={2025},
  volume={39},
  number={4},
  pages={244-253},
  month={Nov.}
  }

@article{huang2024digital,
  title={When digital twin meets generative ai: Intelligent closed-loop network management},
  author={Huang, Xinyu and Yang, Haojun and Zhou, Conghao and He, Mingcheng and Shen, Xuemin and Zhuang, Weihua},
  journal={arXiv preprint arXiv:2404.03025},
  year={2024}
}

@article{yu2024regularized,
  title={Regularized conditional diffusion model for multi-task preference alignment},
  author={Yu, Xudong and Bai, Chenjia and He, Haoran and Wang, Changhong and Li, Xuelong},
  journal={Advances in Neural Information Processing Systems},
  volume={37},
  pages={139968--139996},
  year={2024}
}

@article{emami2024use,
  title={On the use of immersive digital technologies for designing and operating UAVs},
  author={Emami, Yousef and Li, Kai and Almeida, Luis and Ni, Wei},
  journal={arXiv preprint arXiv:2407.16288},
  year={2024}
}

@ARTICLE{liang2024diffusion,
  author={Liang, Ruihuai and Yang, Bo and Chen, Pengyu and Li, Xianjin and Xue, Yifan and Yu, Zhiwen and Cao, Xuelin and Zhang, Yan and Debbah, Mérouane and Vincent Poor, H. and Yuen, Chau},
  journal={IEEE Internet of Things Journal}, 
  title={Diffusion Models as Network Optimizers: Explorations and Analysis}, 
  year={2025},
  volume={12},
  number={10},
  pages={13183-13193},
  month={Jan.}
}

@INPROCEEDINGS{tong2024diffusion,
  author={Tong, Yongju and Kang, Jiawen and Chen, Junlong and Xu, Minrui and Li, Gaolei and Zhang, Weiting and Yan, Xincheng},
  booktitle={GLOBECOM-IEEE Global Communications Conference}, 
  title={Diffusion-based Reinforcement Learning for Dynamic UAV-assisted Vehicle Twins Migration in Vehicular Metaverses}, 
  year={2024},
  volume={},
  number={},
  pages={5156-5161},
  month={Dec.},
  address={Cape Town, South Africa}
  
 }

@ARTICLE{10246260,
  author={Kurunathan, Harrison and Huang, Hailong and Li, Kai and Ni, Wei and Hossain, Ekram},
  journal={IEEE Communications Surveys \& Tutorials}, 
  title={Machine Learning-Aided Operations and Communications of Unmanned Aerial Vehicles: A Contemporary Survey}, 
  year={2024},
  volume={26},
  number={1},
  pages={496-533},
  month={Sep.}}

@ARTICLE{gurses2404digital,
  author = {G{\"u}rses, Anil and
            Reddy, Gautham and
            Masrur, Saad and
            Ozdemir, Ozgur and
            Guven{\c{c}}, {\.I}smail and
            Sichitiu, Mihail L. and
            {\c{S}}ahin, Alphan and
            Alkhateeb, Ahmed and
            Mushi, Magreth and
            Dutta, Rudra},
  journal = {IEEE Communications Magazine},
  title   = {Digital Twins and Testbeds for Supporting AI Research with Autonomous Vehicle Networks},
  year    = {2025},
  volume  = {63},
  number  = {4},
  pages   = {56--62},
  month   = mar
}

@article{CAI2025107064,
title = {DDP-DAR: Network intrusion detection based on denoising diffusion probabilistic model and dual-attention residual network},
journal = {Neural Networks},
volume = {184},
pages = {107064},
year = {2025},
issn = {0893-6080},
author = {Saihua Cai and Yingwei Zhao and Jiaao Lyu and Shengran Wang and Yikai Hu and Mengya Cheng and Guofeng Zhang},
month={Apr.}
}

@INPROCEEDINGS{10610665,
  author={Sridhar, Ajay and Shah, Dhruv and Glossop, Catherine and Levine, Sergey},
  booktitle={IEEE International Conference on Robotics and Automation (ICRA)}, 
  title={NoMaD: Goal Masked Diffusion Policies for Navigation and Exploration}, 
  year={2024},
  volume={},
  number={},
  pages={63-70},
  month={May},
  address={Yokohama, Japan}
  }

@article{10.1145/3626235,
author = {Yang, Ling and Zhang, Zhilong and Song, Yang and Hong, Shenda and Xu, Runsheng and Zhao, Yue and Zhang, Wentao and Cui, Bin and Yang, Ming-Hsuan},
title = {Diffusion Models: A Comprehensive Survey of Methods and Applications},
year = {2023},
issue_date = {April 2024},
publisher = {Association for Computing Machinery},
address = {New York, NY, USA},
volume = {56},
number = {4},
journal = {ACM Comput. Surv.},
month = nov,
articleno = {105},
numpages = {39},
}

@article{zhao2025generative,
  title={Generative AI-enabled wireless communications for robust low-altitude economy networking},
  author={Zhao, Changyuan and Wang, Jiacheng and Zhang, Ruichen and Niyato, Dusit and Sun, Geng and Du, Hongyang and Kim, Dong In and Jamalipour, Abbas},
  journal={arXiv preprint arXiv:2502.18118},
  year={2025}
}

@inproceedings{das2024dronediffusion,
  title={Dronediffusion: Robust quadrotor dynamics learning with diffusion models},
  author={Das, Avirup and Yadav, Rishabh Dev and Sun, Sihao and Sun, Mingfei and Kaski, Samuel and Pan, Wei},
  booktitle={IEEE International Conference on Robotics and Automation (ICRA)},
  pages={1604--1610},
  year={2025},
  organization={IEEE}
}

\end{document}